\documentclass{article} 
\usepackage{iclr2025_ai4na, times}
\usepackage{titlesec}

\usepackage{amsmath,amsfonts,bm}

\def\eqref#1{equation~\ref{#1}}

\def\1{\bm{1}}

\DeclareMathAlphabet{\mathsfit}{\encodingdefault}{\sfdefault}{m}{sl}
\SetMathAlphabet{\mathsfit}{bold}{\encodingdefault}{\sfdefault}{bx}{n}

\usepackage{hyperref}
\usepackage{url}
\usepackage{amsmath, amssymb, bm, graphicx}
\usepackage{booktabs} 
\usepackage{multirow} 
\usepackage{xcolor} 
\usepackage{hyperref} 
\hypersetup{colorlinks=true, linkcolor=blue, citecolor=blue, urlcolor=blue}

\usepackage{adjustbox} 
\usepackage{caption}   
\usepackage{array}
\usepackage{float}
\usepackage{xcolor}    
\usepackage{colortbl}  
\titlespacing{\section}{3pt}{*0.5}{*0.5} 
\titlespacing{\subsection}{0pt}{*0.3}{*0.3} 
\titlespacing{\subsubsection}{0pt}{*0.2}{*0.2} 

\title{MoXGATE: Modality-Aware Cross-Attention for Multi-Omic Gastrointestinal Cancer Subtype Classification}

\author{
Sajib Acharjee Dip\textsuperscript{1}\thanks{Email: sajibacharjeedip@vt.edu}, 
UA Shuvo\textsuperscript{2}, 
D Mallick\textsuperscript{3}, 
Abrar Rahman Abir\textsuperscript{4}, 
Liqing Zhang\textsuperscript{1}\thanks{Corresponding author, Email: lqzhang@cs.vt.edu} \\
\textsuperscript{1}Virginia Tech \\
\textsuperscript{2}University of Dhaka\\
\textsuperscript{3}North South University \\
\textsuperscript{4}Bangladesh University of Engineering and Technology
}

\iclrfinalcopy 
\begin{document}

\maketitle
\begin{abstract}
Cancer subtype classification is crucial for personalized treatment and prognostic assessment. However, effectively integrating multi-omic data remains challenging due to the heterogeneous nature of genomic, epigenomic, and transcriptomic features. In this work, we propose Modality-Aware Cross-Attention MoXGATE, a novel deep-learning framework that leverages cross-attention and learnable modality weights to enhance feature fusion across multiple omics sources. Our approach effectively captures inter-modality dependencies, ensuring robust and interpretable integration. Through experiments on Gastrointestinal Adenocarcinoma (GIAC) and Breast Cancer (BRCA) datasets from TCGA, we demonstrate that MoXGATE outperforms existing methods, achieving 95\% classification accuracy. Ablation studies validate the effectiveness of cross-attention over simple concatenation and highlight the importance of different omics modalities. Moreover, our model generalizes well to unseen cancer types e.g., breast cancer, underscoring its adaptability. Key contributions include (1) a cross-attention-based multi-omic integration framework, (2) modality-weighted fusion for enhanced interpretability, (3) application of focal loss to mitigate data imbalance, and (4) validation across multiple cancer subtypes. Our results indicate that MoXGATE is a promising approach for multi-omic cancer subtype classification, offering improved performance and biological generalizability.
\end{abstract}

\section{Introduction}

Cancer subtyping plays a pivotal role in precision oncology, guiding targeted therapy selection and improving patient outcomes \citep{ahren2009islet}. Gastrointestinal adenocarcinoma (GIAC), a heterogeneous group of malignancies, presents significant challenges in classification due to its high molecular complexity and overlapping subtypes. Traditional histopathological assessments and single-omic biomarkers often fail to capture the full landscape of cancer heterogeneity, underscoring the need for integrative, data-driven approaches.

Recent advances in next-generation sequencing (NGS) technologies have made multi-omic datasets widely available, encompassing gene expression (mRNA), DNA methylation, and miRNA profiles (The Cancer Genome Atlas, 2013) \citep{weinstein2013cancer}. Multi-omic integration enables a comprehensive characterization of tumor biology, improving the robustness of subtyping models. However, effectively leveraging multi-omic data remains an open challenge, as existing approaches struggle with modality heterogeneity, feature redundancy, and computational scalability.

Several recent models have made significant strides in multi-omic cancer subtyping by leveraging advanced deep learning and statistical methodologies, enhancing the ability to integrate diverse biological data for more accurate and robust classification.  moBRCAnet \citep{choi2023mobrca} employs self-attention and simple concatenation, which limits its ability to model inter-modality dependencies, making it suboptimal for capturing feature interactions. DeepMoIC \citep{wu2024deepmoic} applies Graph Convolutional Networks (GCNs) to pan-cancer classification but only identifies three subtypes, making the task inherently easier. It also uses autoencoders for feature extraction, which increases computational overhead. MOGONET \citep{wang2021mogonet} relies heavily on graph construction for each omics modality, and poorly defined graphs fail to capture biological relevance. While it outperforms traditional machine learning models, it remains inferior to attention-based methods. MoGCN \citep{li2022mogcn} utilizes graph-based similarity fusion, but its sensitivity to hyperparameters can impact performance. The graph convolutional model also struggles with scalability and lacks interpretability, despite visualization efforts.

In recent years, multimodal deep learning has emerged as a powerful approach for cancer subtype classification, leveraging diverse multi-omic data such as gene expression, DNA methylation, and miRNA profiles. Advanced models like self-attention encoders \citep{waswani2017attention} and cross-attention mechanisms \citep{wei2020multi} have demonstrated superior feature integration by capturing both within-modality and cross-modality interactions. MMCA \citep{wei2020multi} applies a cross-attention based approach to integrate image and sentence matching. Our approach extends these efforts by incorporating modality-weighted cross-attention \citep{golovanevsky2024one}, allowing adaptive fusion of multi-omic features based on their relative importance. While cross-attention enhances feature alignment, existing limitations include computational complexity, potential modality redundancy, and data imbalance challenges. Addressing these issues with efficient fusion strategies and domain adaptation techniques could further refine multi-omic classification performance.

In this work, we propose a novel attention-based framework for multi-omic integration in cancer subtype classification, specifically focusing on GIAC cancer. These contributions collectively advance the state-of-the-art in cancer subtype classification by improving integration, interpretability, and robustness in multi-omic predictive modeling. Our key contributions are as follows:

\begin{itemize}
    \item Accurate GIAC Cancer Subtyping: To the best of our knowledge, this is the first study that effectively applies attention mechanisms for subtype prediction in GIAC cancer, achieving state-of-the-art performance.
    
    \item Attention and Cross-Attention for Multi-Omic Integration: We employ both self-attention and cross-attention mechanisms to effectively integrate heterogeneous multi-omic data, capturing complex interdependencies between different modalities.
    
    \item Modality-Aware Fusion with Learnable Weights: We introduce learnable modality weights that dynamically adjust the contribution of each omic source, ensuring optimal feature fusion and enhancing predictive performance.
    
    \item Focal Loss for Handling Class Imbalance: Given the high class imbalance in cancer subtyping datasets, we utilize focal loss \citep{ross2017focal} to mitigate the dominance of majority classes, improving classification performance for minority subtypes.
    
    \item Transferability to Other Cancers: Our method is designed to be adaptable to other cancer types beyond GIAC, demonstrating strong generalizability and competitive performance across diverse datasets.
\end{itemize}

\section{Methodology}

\subsection{Model Architecture}
\begin{figure}[ht]
    \centering
    \includegraphics[width=0.95\textwidth]{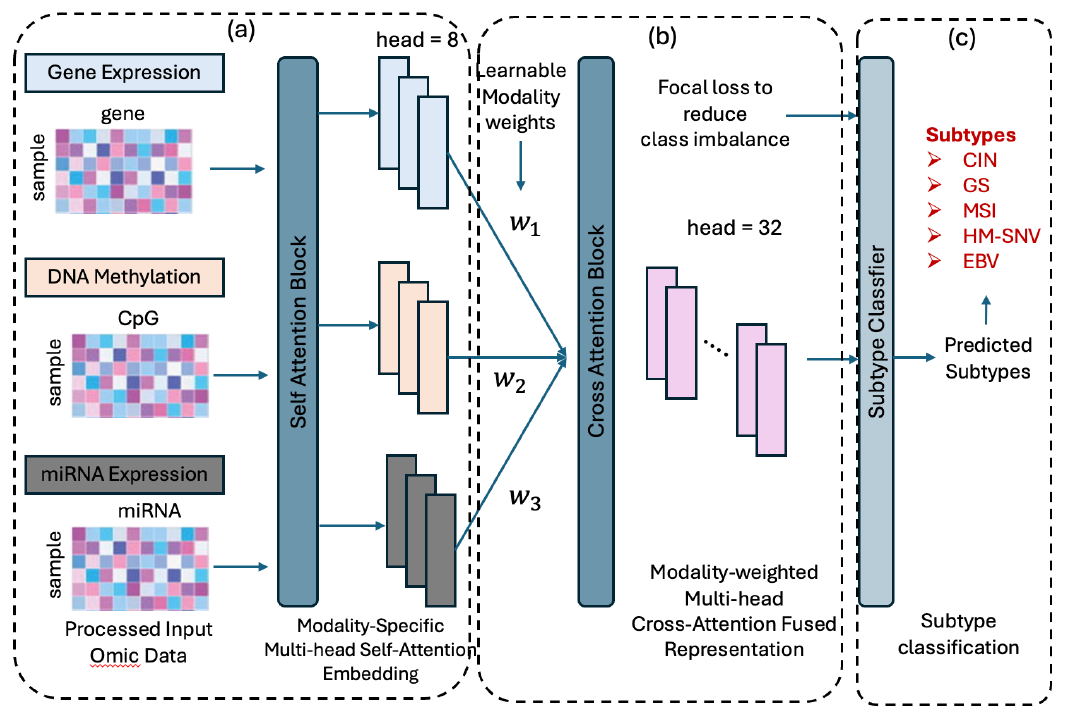}
    \caption{Overview of the Proposed Multi-Omic Cancer Subtype Classification Model. 
    The model consists of three main components: (a) Modality-Specific Self-Attention Encoding, where gene expression, DNA methylation, and miRNA expression undergo independent self-attention transformations to capture high-level feature representations. 
    (b) Modality-Weighted Cross-Attention Fusion, which learns the interdependencies between omic features using learnable modality weights ($w_1, w_2, w_3$) and a multi-head cross-attention mechanism to refine the feature integration. 
    (c) Subtype Classification Module, where the fused representation is passed to a classifier optimized using focal loss, addressing class imbalance in cancer subtyping.}
    
    \label{fig:model_architecture}
\end{figure}

We introduce a novel multi-omics fusion framework that integrates self-attention \citep{waswani2017attention}, modality-weighted cross-attention \citep{wei2020multi}, and focal loss \citep{ross2017focal} optimization. Our approach shown in \ref{fig:model_architecture} optimally encodes high-dimensional multi-omics data, learns the interdependencies among different modalities, and enhances classification through adaptive weighting mechanisms.
\setlength{\textfloatsep}{3pt} 
\setlength{\abovecaptionskip}{1pt} 
\setlength{\belowcaptionskip}{1pt} 

\subsubsection{Self-Attention Encoding for Modality-Specific Representations}

Given an input feature matrix for modality \( m \), defined as:
\begin{equation}
    \mathbf{X}_m = \left[ \mathbf{x}_1, \mathbf{x}_2, ..., \mathbf{x}_N \right]^\top \in \mathbb{R}^{N \times d_m}
\end{equation}
where \( N \) is the number of samples, and \( d_m \) is the feature dimension for modality \( m \), we first apply a linear transformation:
\begin{equation}
    \mathbf{H}_m = \phi \left( \mathbf{X}_m \mathbf{W}_m + \mathbf{b}_m \right) \in \mathbb{R}^{N \times d}
\end{equation}
where \( \mathbf{W}_m \in \mathbb{R}^{d_m \times d} \) and \( \mathbf{b}_m \in \mathbb{R}^{d} \) are trainable parameters, and \( \phi(\cdot) \) denotes a non-linearity (e.g., ReLU).

For the Self-Attention Computation, we compute self-attention by defining:
\begin{align}
    \mathbf{Q}_m &= \mathbf{H}_m \mathbf{W}_Q, \quad
    \mathbf{K}_m = \mathbf{H}_m \mathbf{W}_K, \quad
    \mathbf{V}_m = \mathbf{H}_m \mathbf{W}_V \\
    \mathbf{A}_m &= \text{softmax} \left( \frac{\mathbf{Q}_m \mathbf{K}_m^\top}{\sqrt{d}} \right) \\
    \mathbf{Z}_m &= \mathbf{A}_m \mathbf{V}_m + \mathbf{H}_m
\end{align}
where \( \mathbf{W}_Q, \mathbf{W}_K, \mathbf{W}_V \in \mathbb{R}^{d \times d} \) are trainable weight matrices.

\subsubsection{Modality-Weighted Cross-Attention Fusion}
We integrate different omics sources via a modality-weighted cross-attention mechanism. Let \( \mathbf{Z}_1, \mathbf{Z}_2, \mathbf{Z}_3 \) be self-attention outputs for three modalities (Gene, Methylation, miRNA). We first construct a stacked representation:
\begin{equation}
    \mathbf{C} = \left[ \mathbf{Z}_1, \mathbf{Z}_2, \mathbf{Z}_3 \right] \in \mathbb{R}^{3 \times N \times d}
\end{equation}

For the Cross-Modality Attention, We compute the query, key, and value projections:
\begin{align}
    \mathbf{Q}_c &= \mathbf{C} \mathbf{W}_Q^c, \quad
    \mathbf{K}_c = \mathbf{C} \mathbf{W}_K^c, \quad
    \mathbf{V}_c = \mathbf{C} \mathbf{W}_V^c
\end{align}
where \( \mathbf{W}_Q^c, \mathbf{W}_K^c, \mathbf{W}_V^c \in \mathbb{R}^{d \times d} \) are trainable.

The cross-attention weights are:
\begin{equation}
    \mathbf{A}_c = \text{softmax} \left( \frac{\mathbf{Q}_c \mathbf{K}_c^\top}{\sqrt{d}} \right)
\end{equation}

The fused representation is:
\begin{equation}
    \mathbf{F} = \mathbf{A}_c \mathbf{V}_c
\end{equation}

\subsubsection{Modality Importance Learning}
To ensure balanced fusion, we introduce learnable modality weights \( \mathbf{w} \):
\begin{equation}
    \mathbf{w} = \left[ w_1, w_2, w_3 \right] \quad \text{where} \quad \sum_{i=1}^{3} w_i = 1
\end{equation}
Initially, the modality weights \( w_1, w_2, w_3 \) were uniformly set to 0.33 each, reflecting equal contribution from all omics modalities at the start of training.

The final weighted fusion output is:
\begin{equation}
    \mathbf{F}^{\text{final}} = w_1 \mathbf{F}_1 + w_2 \mathbf{F}_2 + w_3 \mathbf{F}_3
\end{equation}

\subsubsection{Classification with Focal Loss}
The fused representation is classified via:
\begin{equation}
    \hat{\mathbf{y}} = \sigma \left( \mathbf{W}_f \mathbf{F}^{\text{final}} + \mathbf{b}_f \right)
\end{equation}
where \( \mathbf{W}_f \in \mathbb{R}^{d \times K} \) is the weight matrix, and \( K \) is the number of classes.

To handle class imbalance, we employ Focal Loss:
\begin{equation}
    \mathcal{L}_{\text{focal}} = -\sum_{i=1}^{K} \alpha_i \left(1 - p_i\right)^\gamma y_i \log p_i
\end{equation}
where \( p_i \) is the predicted probability for class \( i \), \( \alpha_i \) is the class-specific weighting, \( \gamma \) is the focusing parameter. We set \( \alpha_i = 1\) and \( \gamma = 2\) for all experiments.

\subsubsection{Overall Optimization and Training}
The final optimization objective consists of:
\begin{equation}
    \mathcal{L} = \mathcal{L}_{\text{focal}} + \lambda_1 \|\mathbf{w} - \mathbf{1}\|^2 + \lambda_2 \|\mathbf{W}_c\|_F^2
\end{equation}
where \( \lambda_1 \) ensures modality weights remain balanced and \( \lambda_2 \) applies Frobenius norm regularization to prevent overfitting.

\subsubsection{Experimental Setup}  
We used modality-specific multi-head self-attention encoders with 8 heads and a dropout rate of 0.1, without sharing weights across modalities. The outputs from each modality were concatenated and passed through a cross-attention layer to model inter-modality interactions, with the combined embedding dimension set to 256 and 32 attention heads. The classifier uses ReLU activation and a dropout rate of 0.3. The final embedding layer has a dimensionality of 128, followed by a fully connected layer mapping to the number of output classes. We trained the model using the AdamW optimizer with a learning rate of \(10^{-4}\) and a weight decay of \(10^{-2}\).

This framework integrates multi-omics data through dedicated self-attention encoders, modality-weighted cross-attention, and focal loss-based classification, enabling robust and effective multi-omics feature fusion.

\section{Results}
\subsection{Performance Comparison on GIAC Subtype Classification}

The results in Table \ref{tab:model_comparison} confirm that our model achieves the highest accuracy (0.95) and F1-score (0.94), outperforming all baselines. AE + Cross Attn struggles with feature integration (accuracy 0.72, F1-score 0.79), while Self Attn + Gated Attn improves feature extraction (accuracy 0.86) but lacks cross-modal fusion. Self Attn + Mod Gated Attn and moBRCAnet (accuracy 0.93, F1-score 0.92) perform competitively but fail to explicitly model inter-omic dependencies.

Our modality-aware cross-attention with weighted fusion effectively balances omic contributions, surpassing simple concatenation and self-attention methods. The results highlight the importance of learning cross-modal interactions for robust subtype classification, demonstrating the superiority of our approach in integrating multi-omic data.

\setlength{\textfloatsep}{3pt} 
\setlength{\abovecaptionskip}{1pt} 
\setlength{\belowcaptionskip}{1pt} 
\begin{table}[ht]
    \centering
    \renewcommand{\arraystretch}{1.2}  
    \setlength{\tabcolsep}{8pt}  
    \caption{Comparison of Model Performance}
    \label{tab:model_comparison}
    \begin{adjustbox}{max width=\textwidth}
    \begin{tabular}{lcccc}
        \toprule
        \textbf{Model} & \textbf{Accuracy} & \textbf{Precision} & \textbf{Recall} & \textbf{F1-Score} \\
        \midrule
        AE + Cross Attn       & 0.72 & 0.93 & 0.72 & 0.79 \\
        Self Attn + Gated Attn & 0.86 & 0.92 & 0.86 & 0.89 \\
        Self Attn + Mod Gated Attn & 0.93 & 0.94 & 0.92 & 0.92 \\
        moBRCA-net             & 0.93 & 0.94 & 0.92 & 0.92 \\
        \rowcolor{gray!20} \textbf{Ours} & \textbf{0.95} & \textbf{0.96} & \textbf{0.95} & \textbf{0.94} \\
        \bottomrule
    \end{tabular}
    \end{adjustbox}
\end{table}

\subsection{Importance of Different Omics for Final Prediction}

The results in Table \ref{tab:ablation_study} illustrate the contribution of each omic modality to the final classification performance. Among the individual omics, methylation data provides the highest accuracy (0.95) and F1-score (0.94), indicating its strong discriminative power for cancer subtyping. Gene expression data follows closely, achieving an accuracy of 0.91, suggesting that gene-level variations contribute significantly to subtype differentiation. However, miRNA alone shows the lowest accuracy (0.87) and F1-score (0.90), highlighting its limited capability to capture subtype-specific variations when used in isolation.

When combining two modalities, Methylation + Gene achieves an accuracy of 0.94, demonstrating a synergistic effect in feature fusion. Similarly, Methylation + miRNA also achieves 0.95 accuracy, suggesting that methylation patterns enhance the discriminative power of miRNA features. In contrast, the combination of Gene + miRNA yields a lower accuracy of 0.85, reinforcing the idea that without methylation data, the model struggles to generalize well.

\setlength{\textfloatsep}{2pt} 
\setlength{\abovecaptionskip}{1pt} 
\setlength{\belowcaptionskip}{1pt} 
\begin{table}[ht]
    \centering
    \renewcommand{\arraystretch}{1.2}  
    \setlength{\tabcolsep}{8pt}  
    \caption{Performance Comparison Across Different Omics Modalities}
    \label{tab:ablation_study}
    \begin{adjustbox}{max width=\textwidth}
    \begin{tabular}{lcccc}
        \toprule
        \textbf{Modality} & \textbf{Accuracy} & \textbf{Precision} & \textbf{Recall} & \textbf{F1 Score} \\
        \midrule
        Gene               & 0.91  & 0.95  & 0.91  & 0.92  \\
        Methylation        & 0.95  & 0.96  & 0.95  & 0.94  \\
        miRNA              & 0.87  & 0.95  & 0.87  & 0.90  \\
        Gene + Methylation & 0.94  & 0.96  & 0.94  & 0.94  \\
        Gene + miRNA       & 0.85  & 0.95  & 0.85  & 0.88  \\
        Methylation + miRNA& 0.95  & 0.96  & 0.95  & 0.94  \\
        \rowcolor{gray!20} \textbf{All (Combined)} & \textbf{0.95}  & \textbf{0.96}  & \textbf{0.95}  & \textbf{0.94}  \\
        \bottomrule
    \end{tabular}
    \end{adjustbox}
\end{table}

The best-performing model is the fully integrated multi-omic approach, which incorporates gene expression, methylation, and miRNA. This results in an accuracy of 0.95 and an F1-score of 0.94, confirming that a comprehensive multi-omic fusion is essential for optimal cancer subtype classification. The observed performance gains highlight the importance of cross-modal interactions and justify our use of a modality-aware cross-attention framework to leverage the complementary strengths of each omic type.

\subsection{Performance on Other Cancer Data}
\setlength{\textfloatsep}{2pt} 
\setlength{\abovecaptionskip}{1pt} 
\setlength{\belowcaptionskip}{1pt} 
\begin{table}[ht]
\centering
\renewcommand{\arraystretch}{1.2}
\begin{tabular}{lcc}
    \toprule
    \textbf{Model} & \textbf{Accuracy} & \textbf{F1-Score} \\
    \midrule
    AE+Cross Attn & 0.82 & 0.79 \\
    moBRCANet & 0.87 & 0.86 \\
    \rowcolor{gray!20} \textbf{Ours} & \textbf{0.89} & \textbf{0.88} \\
    \bottomrule
\end{tabular}
\caption{Performance comparison on breast cancer subtype classification. Our model achieves the best performance, demonstrating strong generalization.}
\label{tab:breast_cancer_results}
\end{table}
To further validate the generalizability of our method, we conducted experiments on the TCGA-BRCA dataset shown in Table \ref{tab:breast_cancer_results}, which consists of 1,057 breast cancer samples. The dataset includes five intrinsic subtypes from the PAM50 classification: luminal A, luminal B, HER2 overexpression, basal-like, and normal-like cancers. We followed the same preprocessing steps as applied to the GIAC dataset, ensuring consistency across experiments. The dataset was split into 80\% training and 20\% testing, with 10\% of the training data used for validation.

As shown in Table 3, our model achieves an accuracy of 0.89 and an F1-score of 0.88, outperforming existing approaches such as AE+Cross Attention (0.82 accuracy) and moBRCANet (0.87 accuracy). These results demonstrate that our modality-aware cross-attention approach effectively generalizes across different cancer types, reinforcing its robustness in multi-omic cancer subtype classification.


\bibliography{iclr2025_ai4na}
\bibliographystyle{iclr2025_conference}

\appendix
\section{Appendix}
\subsection{Dataset}

\subsection{GIAC Cancer and Subtypes}
Gastrointestinal Adenocarcinomas (GIACs) include four major cancer types: Colon Adenocarcinoma (COAD), Rectum Adenocarcinoma (READ), Stomach Adenocarcinoma (STAD), and Esophageal Carcinoma (ESCA). These cancers exhibit distinct histopathological and molecular characteristics:

\begin{itemize}
    \item \textbf{COAD (Colon Adenocarcinoma):} A common gastrointestinal malignancy characterized by chromosomal instability (CIN) and microsatellite instability (MSI), with additional classifications based on molecular features.
    \item \textbf{READ (Rectum Adenocarcinoma):} Similar to COAD but arises in the rectum, sharing molecular features but influenced by distinct anatomic and treatment considerations.
    \item \textbf{STAD (Stomach Adenocarcinoma):} A highly heterogeneous cancer associated with multiple subtypes, including Epstein-Barr virus (EBV)-associated tumors, MSI-high tumors, and genomically stable (GS) subtypes.
    \item \textbf{ESCA (Esophageal Carcinoma):} A rare but aggressive cancer exhibiting CIN and MSI features, often linked to environmental and genetic risk factors.
\end{itemize}

\subsection{Dataset Statistics}
The dataset used in this study is sourced from The Cancer Genome Atlas (TCGA) \citep{weinstein2013cancer}, containing multi-omic profiles for GIAC cancers. We specifically focus on molecular subtyping based on genetic and epigenetic alterations. The dataset includes the following samples:

\begin{table}[ht]
\centering
\renewcommand{\arraystretch}{1.2}
\setlength{\tabcolsep}{2pt} 
\begin{tabular}{lcccc} 
    \toprule
    \textbf{Abbreviation} & \textbf{Study Name} & \textbf{Subtype} \\ 
    & & \textbf{Classification} & \textbf{Subtypes} & \textbf{Samples} \\ 
    \midrule
    COAD & Colon Adenocarcinoma & Molecular & CIN, GS, MSI, HM-SNV, EBV & 341 \\
    ESCA & Esophageal Carcinoma & Molecular & CIN, GS, MSI, HM-SNV, EBV & 79 \\
    READ & Rectum Adenocarcinoma & Molecular & CIN, GS, MSI, HM-SNV, EBV & 118 \\
    STAD & Stomach Adenocarcinoma & Molecular & CIN, GS, MSI, HM-SNV, EBV & 383 \\
    \bottomrule
\end{tabular}
\caption{GIAC Cancer Subtypes and Sample Distribution from TCGA. The four studied cancers include Colon Adenocarcinoma (COAD), Esophageal Carcinoma (ESCA), Rectum Adenocarcinoma (READ), and Stomach Adenocarcinoma (STAD), with five molecular subtypes.}
\label{tab:giac_data}
\end{table}

\subsection{Molecular Subtypes in GIACs}
Molecular subtyping in GIACs has been extensively studied using gene expression, oncogenic pathways, and histopathological criteria. However, traditional clustering approaches often struggle with the biological complexity inherent to these cancers. Our study leverages genomic, epigenomic, and transcriptomic data to define robust molecular subtypes. \citep{liu2018comparative}

\textbf{Key Subtype Characteristics:}
\begin{itemize}
    \item \textbf{EBV+ (Epstein-Barr Virus Positive):} Predominantly found in stomach cancers, characterized by extensive DNA hypermethylation.
    \item \textbf{MSI (Microsatellite Instability):} Associated with defective DNA mismatch repair, leading to a high mutation burden.
    \item \textbf{HM-SNV(Hypermutated-Single Nucleotide Variants):} Defined by an SNV-predominant mutation profile, often linked to POLE mutations.
    \item \textbf{CIN (Chromosomal Instability):} Characterized by large-scale chromosomal alterations, frequently found in GIAC tumors.
    \item \textbf{GS (Genome Stable):} Lacks significant chromosomal aberrations, representing a smaller but distinct subset of tumors.
\end{itemize}

The dataset integrates multiple molecular modalities, including mutation profiles, copy-number variations, and DNA methylation, ensuring a comprehensive framework for subtype classification.

\section{Data Processing Pipeline}

To ensure a robust and unbiased evaluation, we utilized three cancer datasets (COAD, READ, STAD) for training and validation, while reserving the ESCA dataset exclusively for testing. Each cancer type in our dataset is categorized into five molecular subtypes. We performed a 90-10 split on the training dataset, where 90\% of the samples were used for model training, and 10\% for validation.

For feature preprocessing, we applied a two-step missing value handling strategy. First, we eliminated features with more than 40\% missing values to ensure data reliability. Second, for the remaining missing values, we applied median imputation, filling in missing entries with the median value of the respective feature.

To maintain biological consistency across datasets, we selected only features that were common across all four cancer types. This yielded the following shared features:
\begin{itemize}
    \item \textbf{Common Gene Expression Features:} 20,530
    \item \textbf{Common DNA Methylation Features:} 23,381
    \item \textbf{Common miRNA Features:} 746
\end{itemize}

Following this preprocessing, our \textbf{final dataset} consisted of:
\begin{itemize}
    \item \textbf{Training and validation set:} 842 samples
    \item \textbf{Test set (ESCA):} 79 samples  
    \item \textbf{Final train-validation split:} 757 training samples and 85 validation samples
\end{itemize}

This data processing pipeline ensures that the model is trained on a diverse set of cancers while testing on a separate cancer type, providing a realistic evaluation of model generalizability across GIAC subtypes.

\section{Ablation Study}
\subsection{Ablation Study of Attention head}
The number of heads in a multi-head cross-attention layer plays a critical role in capturing diverse feature interactions across omics modalities. The ablation study, as presented in Table \ref{tab:ablation_heads}, evaluates the performance of our model with 8, 16, and 32 attention heads. The results indicate that increasing the number of heads from 8 to 16 does not significantly impact performance, maintaining an accuracy of 94\%. However, when the number of heads is increased to 32, the model achieves a slight improvement, reaching the highest accuracy of 95\% along with a higher recall (0.95) and precision (0.96).

This improvement suggests that with a greater number of heads, the model is able to attend to finer-grained relationships among multi-omic features, thereby improving its ability to extract meaningful subtype-specific patterns. However, while a larger number of heads provides marginal gains, further increasing this number may introduce computational overhead without substantial performance benefits. Thus, 32 heads was selected as the optimal configuration, balancing both accuracy and computational efficiency.
\begin{table}[ht]
\centering
\renewcommand{\arraystretch}{1.2}
\setlength{\tabcolsep}{8pt} 
\begin{tabular}{c c c c c} 
    \toprule
    \textbf{Heads} & \textbf{Accuracy} & \textbf{Precision} & \textbf{Recall} & \textbf{F1-Score} \\ 
    \midrule
    8  & 0.94 & 0.96 & 0.94 & 0.94 \\
    16 & 0.94 & 0.96 & 0.94 & 0.94 \\
    \rowcolor{gray!20} 
    32 & 0.95 & 0.96 & 0.95 & 0.94 \\
    \bottomrule
\end{tabular}
\caption{Ablation study on the effect of different numbers of heads in the cross-attention layer. The best-performing setting is highlighted.}
\label{tab:ablation_heads}
\end{table}

\subsection{Ablation Study on Model Architectural Components}

To further evaluate the impact of different architectural choices, we conducted an ablation study by introducing Batch Normalization (BatchNorm), Skip Connections, and Feedforward Attention separately and in combination. Table \ref{tab:ablation_components} summarizes the performance of these variations.

The model with BatchNorm achieved the lowest accuracy (0.89) and recall (0.90), indicating that normalizing intermediate layers did not contribute positively to performance, possibly due to the already normalized omics data. Adding Skip Connections improved the model accuracy to 0.92, showing that residual connections help preserve gradient flow and avoid vanishing gradients. The Feedforward Attention mechanism resulted in the highest accuracy (0.94) and F1-score (0.95), suggesting that additional transformation layers improve feature extraction. However, combining both Skip Connections and Feedforward Attention did not yield further improvements, stabilizing at an accuracy of 0.92.

These findings indicate that while Feedforward Attention enhances feature interactions, BatchNorm is not beneficial for this problem, and Skip Connections alone do not provide significant improvements. As a result, we excluded BatchNorm and retained Feedforward Attention without additional Skip Connections in our final model to achieve optimal performance.
\begin{table}[ht]
\centering
\renewcommand{\arraystretch}{1.2}
\setlength{\tabcolsep}{8pt} 
\begin{tabular}{l c c c c} 
    \toprule
    \textbf{Ablation} & \textbf{Accuracy} & \textbf{Precision} & \textbf{Recall} & \textbf{F1-Score} \\ 
    \midrule
    w/ BatchNorm                  & 0.89 & 0.95 & 0.90 & 0.91 \\
    w/ Skip Connection            & 0.92 & 0.95 & 0.92 & 0.93 \\
    w/ Feedforward Attention      & 0.94 & 0.95 & 0.93 & 0.95 \\
    \rowcolor{gray!20} 
    w/ Skip + Feedforward Attn    & 0.92 & 0.95 & 0.92 & 0.92 \\
    \bottomrule
\end{tabular}
\caption{Ablation study on different model components. The last row highlights the combination of skip connection and feedforward attention, which did not provide additional performance gains.}
\label{tab:ablation_components}
\end{table}

\section{Discussions}
Our proposed Modality-Aware Cross-Attention model demonstrates state-of-the-art performance for multi-omic cancer subtype classification, effectively integrating heterogeneous omics data sources. The cross-attention mechanism, combined with learnable modality weights, enhances the fusion of gene expression, DNA methylation, and miRNA data, capturing intricate inter-modality dependencies. The ablation studies confirm that cross-attention outperforms simple concatenation, emphasizing its significance in multi-omic integration. Additionally, the results highlight the dominance of methylation and gene expression data in driving classification performance, aligning with biological insights into cancer heterogeneity. The strong generalization to breast cancer subtypes further underscores the robustness and transferability of our approach beyond gastrointestinal adenocarcinoma (GIAC).

Despite these advancements, certain limitations persist. First, while cross-attention improves modality fusion, it inherently increases computational complexity, making it less scalable for ultra-large datasets. Additionally, although modality weights provide insight into the relative importance of omics data, they do not explicitly model dynamic feature importance at the patient level, potentially limiting interpretability for individualized cancer profiling. Future work should explore efficient self-attention mechanisms to reduce complexity and incorporate patient-specific attention weighting for improved personalization.

\section{Acknowledgements}
We gratefully acknowledge the support from the Department of Computer Science at Virginia Tech for providing computational resources essential to this work. This research was partially supported by funding from the National Science Foundation (NSF). We also thank the members of the Zhang Lab for their valuable feedback, insights, and discussions throughout the development of this project.

\end{document}